\documentclass[fleqn,10pt]{wlscirep}
\title{LLMs are Introvert}

\author[2]{Litian Zhang}
\author[1$\dagger$]{Xiaoming Zhang}
\author[1]{Bingyu Yan}
\author[1]{Ziyi Zhou}
\author[3]{Bo Zhang}
\author[1]{Zhenyu Guan}
\author[2]{Xi Zhang}
\author[2]{Chaozhuo Li}

\affil[1]{Beihang University, Beijing 100191, China}
\affil[2]{Beijing
University of Posts and Telecommunications, Beijing 100876, China}
\affil[3]{Nanjing Normal University}

\affil[$\dagger$]{Correspondence: yolixs@buaa.edu.cn}

\usepackage{CJKutf8}
\usepackage{bm}
\usepackage{hyperref}
\usepackage{xcolor}
\usepackage{verbatim}
\usepackage{booktabs} 
\usepackage{algorithm}
\usepackage{algorithmic}
\usepackage{epsfig}
\usepackage{color}
\usepackage{amsmath}
\usepackage{graphicx,subfigure}
\usepackage{multirow}
\usepackage{makecell}
\usepackage{float}
\usepackage{microtype}
\usepackage{graphicx} 
\usepackage{hyperref}
\usepackage{makecell}
\usepackage{amsmath}
\usepackage{hyperref}

\usepackage{multirow}
\usepackage{subfigure}

\usepackage{algorithm}
\usepackage{algorithmic}

\usepackage{balance}
\usepackage{booktabs}
\usepackage{url}
\usepackage{color}
\usepackage{colortbl}
\usepackage{xcolor}
\usepackage{xspace}
\usepackage{adjustbox}
\usepackage{booktabs}
\usepackage{tabularx}
\usepackage{array} 
\usepackage{graphicx}
\usepackage{subcaption}

\usepackage{amssymb}
\usepackage[most]{tcolorbox} 
\tcbuselibrary{theorems} 

\usepackage{threeparttablex}

\newtcolorbox{SIP}[1]{
  colback=blue!5,
  colframe=blue!35!black,
  fonttitle=\bfseries,
  title={The prompt template for the SIP-enhance thinking process},
}

\newtcolorbox{Action}[1]{
  colback=blue!5,
  colframe=blue!35!black,
  fonttitle=\bfseries,
  title={The prompt template for the agent action process},
}

\newtcolorbox{micro}[1]{
  colback=blue!5,
  colframe=blue!35!black,
  fonttitle=\bfseries,
  title={The prompt template for the micro-action process},
}
\newtcolorbox{code}[1]{
  colback=blue!5,
  colframe=blue!35!black,
  fonttitle=\bfseries,
  title={The prompt template for the code QA task},
}
\newtcolorbox{web}[1]{
  colback=blue!5,
  colframe=blue!35!black,
  fonttitle=\bfseries,
  title={The prompt template for the web QA task},
}
\newtcolorbox{tableqa}[1]{
  colback=blue!5,
  colframe=blue!35!black,
  fonttitle=\bfseries,
  title={The prompt template for the table QA task},
}
\newtcolorbox{summarization}[1]{
  colback=blue!5,
  colframe=blue!35!black,
  fonttitle=\bfseries,
  title={The prompt template for the summarization task},
}

\newtcolorbox{user}[1]{
  colback=blue!5,
  colframe=blue!35!black,
  fonttitle=\bfseries,
  title={User Query},
}
\newtcolorbox{gpt4}[1]{
  colback=brown!5,
  colframe=brown!35!black,
  fonttitle=\bfseries,
  title={GPT-4},
}
\newtcolorbox{gpt35}[1]{
  colback=green!5,
  colframe=green!35!black,
  fonttitle=\bfseries,
  title={GPT-3.5-Turbo},
}
\newtcolorbox{vicuna7b}[1]{
  colback=gray!5,
  colframe=gray!95!black,
  fonttitle=\bfseries,
  title={Vicuna-7B},
}
\newtcolorbox{vicuna13b}[1]{
  colback=purple!5,
  colframe=purple!35!black,
  fonttitle=\bfseries,
  title={Vicuna-13B},
}

\newtcolorbox{mt-gpt4}[1]{
  colback=brown!5,
  colframe=brown!35!black,
  fonttitle=\bfseries,
  title={GPT-4 Defended by Multi-turn Dialogue Defense},
}
\newtcolorbox{incontext-gpt4}[1]{
  colback=brown!5,
  colframe=brown!35!black,
  fonttitle=\bfseries,
  title={GPT-4 Defended by In-context Learning Defense},
}

\newtcolorbox{mt-gpt35}[1]{
  colback=green!5,
  colframe=green!35!black,
  fonttitle=\bfseries,
  title={GPT-3.5-turbo Defended by Multi-turn Dialogue Defense},
}
\newtcolorbox{incontext-gpt35}[1]{
  colback=green!5,
  colframe=green!35!black,
  fonttitle=\bfseries,
  title={GPT-3.5-turbo Defended by In-context Learning Defense},
}

\newtcolorbox{vicuna7b-white}[1]{
  colback=gray!5,
  colframe=gray!95!black,
  fonttitle=\bfseries,
  title={Vicuna-7B Defended by White-box Defense with GPT-4 Responses},
}
\newtcolorbox{vicuna13b-white}[1]{
  colback=purple!5,
  colframe=purple!35!black,
  fonttitle=\bfseries,
  title={Vicuna-13B Defended by White-box Defense with GPT-4 Responses}
}
\newtcolorbox{incontext}[1]{
  colback=purple!5,
  colframe=purple!35!black,
  fonttitle=\bfseries,
  title={GPT-3.5-turbo Defended by the Black-box Defense with In-context Learnig},
}

  \usepackage{subfigure}

\begin{abstract}
The exponential growth of social media platforms and generative AI technologies has revolutionized the creation and dissemination of multimedia information, fostering connectivity but also facilitating the rapid spread of misinformation and harmful content. Understanding the dynamics of information propagation and developing control strategies are vital for mitigating these adverse effects while promoting accurate information dissemination.
Traditional models, such as the SIR (Susceptible-Infected-Recovered) framework, offer foundational insights but fall short in capturing the complexities of online social media. Advanced methods like attention-based algorithms and graph neural networks have improved the accuracy of simulations but often neglect user behavior and psychological changes.
Large language models (LLMs), with their human-like reasoning and cognitive abilities, present a novel opportunity for modeling the psychological aspects of information propagation. Our simulation environment using LLMs records agents' attitudes, emotions, behaviors, and their responses to information over time. Initial results reveal a considerable gap between LLM-generated behaviors and actual human behavior, particularly in stance detection and psychological state alignment.
To bridge this gap, we conducted a detailed evaluation based on Social Information Processing Theory, which revealed significant discrepancies in goal setting and feedback evaluation between LLMs and humans. These differences are attributed to the LLM training process, which lacks the personalized emotional processing essential for human communication.
To address these shortcomings, we propose a Social Information Processing-based Chain of Thought (SIP-CoT) mechanism, supplemented by emotion-guided memory. This approach enhances the encoding and interpretation of social cues, the establishment of personalized goals, and feedback evaluation. Experimental results show that SIP-CoT-enhanced LLM agents process social information more effectively, exhibiting more human-like attitudes, behaviors, and emotions.
In summary, our research identifies key gaps in LLM-based simulations of information propagation and demonstrates that integrating SIP-CoT mechanisms and emotion-guided memory can significantly improve the social intelligence of LLM agents, bringing their behavior closer to that of real humans in dissemination scenarios.

\end{abstract}
\begin{document}
\begin{CJK*}{UTF8}{gbsn}
\flushbottom
\maketitle
%
%
\clearpage
\begin{figure*}[!ht]
  \centering
  \includegraphics[width=1.0\textwidth]{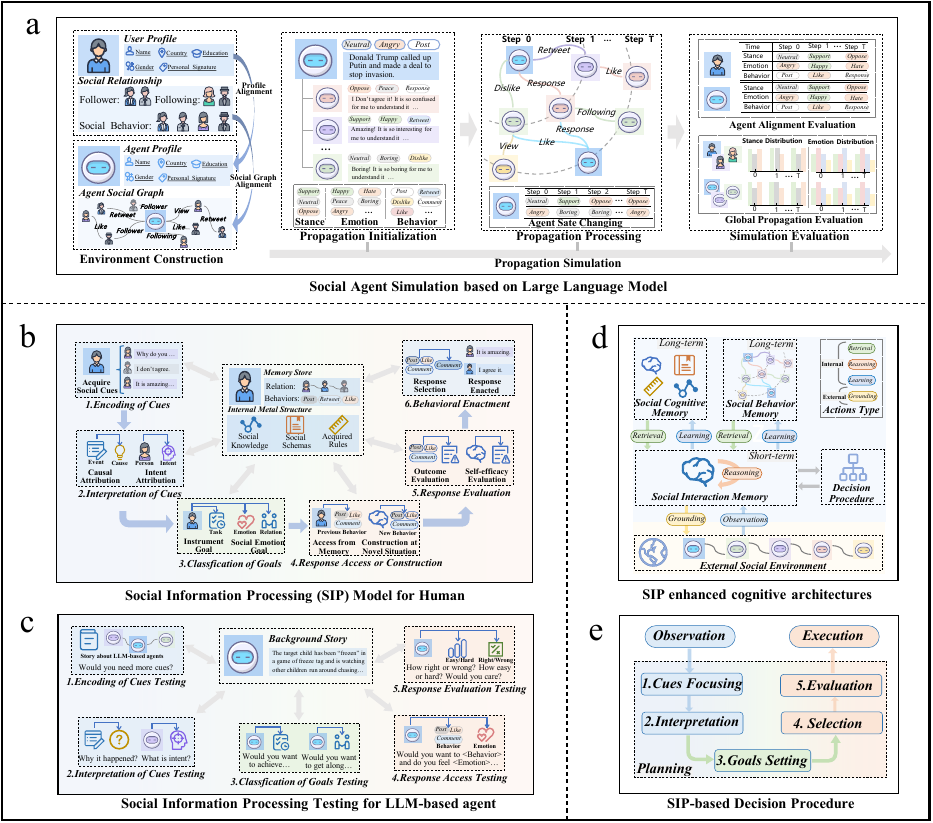}
  \caption{\textbf{Overview of the SIP-enhanced social-agent framework.}
\textbf{a}, Time-step social-agent simulation driven by LLMs, comprising environment construction, simulation initialization, interaction processing and two-tier evaluation (agent alignment and global propagation). 
b, Six-stage Social Information Processing (SIP) model delineating human social cognition. 
c, SIP testing paradigm for LLM agents that assesses encoding, interpretation, goal classification, response access and response evaluation via Likert-scaled tasks. 
\textbf{d}, Proposed cognitive architecture that embeds SIP stages, long- and short-term social memory and decision-making modules within an LLM agent. 
\textbf{e}, Resulting SIP-based decision procedure integrating observation, planning and execution phases.}
  \label{fig:example}
\end{figure*}


Large language models (LLMs) \cite{openai2022chatgpt, chowdhery2023palm, openai2023gpt4, anthropic2023claude} have achieved remarkable success across a variety of natural language processing tasks, including machine translation \cite{zhang2023prompting,wang2023document,zhang2023prompting}, summarization \cite{zhang2024systematic,van2024adapted,ma2024large}, and question-answering \cite{shao2023prompting,singhal2025toward,kamalloo2023evaluating}. 
Their advanced capabilities have garnered significant attention from both academia and industry \cite{liu2023summary,meyer2023chatgpt}. 
Despite these linguistic task advancements, current LLMs remain markedly introvert, which failing to interpret subtle social cues as  with human‑level nuance\cite{10.1162/opmi_a_00160}. 
These shortcomings raise concerns about their effectiveness in applications requiring social cognition, such as LLM‑driven social simulations and human-agent interactions\cite{park2023generative}. 
Addressing these challenges is crucial for developing LLMs systems capable of engaging in realistic social environments.


Social simulations, grounded in agent‑based computational modelling\cite{epstein2012generative}, enabling researchers to study complex interactions and emergent behaviors that arise from individual agents following simple rules\cite{janssen2006empirically, crooks2011introduction}.
Integrating LLMs into simulations offers a paradigm shift from static rule-based architectures toward data-driven emergent linguistic reasoning\cite{zeng2024exploring, gao2024large}, reducing the barrier for  constructing rich simulations while enabling rapid adaptability across dynamic interaction contexts.
However, current LLM-driven simulations fail to capture nuanced social cognition processes essential for realistic interactions \cite{sap2022neural, huang2024social}. 
Inaccurate modelling of individual perception cascades into systematic network‑wide distortions, compromising the accuracy of large-scale social dynamic modeling.


To systematically evaluate the performance of LLMs in social simulations, comprehensive simulations employing LLM-based agents are designed to replicate human social dynamics within controlled environments (Figure~1a). 
By leveraging the real-world interaction data\cite{medvedev2017anatomy}, the simulations aim to assess the ability of LLM agents to perform with in social platform. 
The simulation process involves multiple agents interacting over simulated time steps, enabling the examination of both macro-level social propagation and micro-level individual behaviors. However, the results reveal that LLM-based agents often struggle with interpreting tiny social cues and generating formulaic responses, leading to simulations that lack realism and depth (Figure~1a). 
This highlights the importance in accurately simulating human-like social behaviors, underscoring the need for enhanced cognitive architectures that better mimic human social cognition.

The Social Information Processing (SIP) model \cite{dodge1986social,huesmann1988social} provides a cognitive-level framework for analyzing human social interactions, particularly addressing current limitations through its structured processing stages (Figure~1b).
The SIP model delineates a series of cognitive processes, including cue encoding, interpretation, goal clarification, response generation, and decision-making \cite{crick1994role}. 
When evaluated through the SIP testing, LLM-based agents exhibit rigid and less nuanced processing of social information across the critical stages of social cognition as human (Figure~1c). 
Specifically, the agents struggle with interpreting ambiguous social cues and fail to display appropriate emotional responses, such as feelings of rejection or anger, which are essential components of human social interactions. 
In the stages of goal classification, LLMs often default to generic or overly formal behaviors, lacking the diversity observed in human responses. 
Additionally, during response evaluation, the agents show limited ability to assess the moral and social implications of potential actions, indicating a gap in their decision-making processes. 
These systematic deficiencies across multiple stages of the SIP framework underscore the substantial challenges in developing LLM architectures capable of emulating the flexible, context-sensitive social cognition that characterizes human interaction.



{Addressing these constraints,} we develop an SIP-enhanced cognitive architecture for LLM-based agents that achieves closer alignment with human social information processing mechanisms (Figures~1d and 1e).
This architecture integrates advanced memory models, including both long-term and short-term social memories, to store and retrieve social knowledge and past interactions. 
It also incorporates a decision-making process informed by social cues and goals, enabling agents to interpret social stimuli, set context-appropriate objectives, and evaluate potential responses.
Experimental evaluations demonstrate that agents equipped with this architecture exhibit significantly improved performance in social simulations, displaying more human-like patterns in cue interpretation, goal formulation, and response selection. 
Incorporating social psychological frameworks into AI development significantly enhances LLM agents' social capabilities\cite{}, advancing both simulation realism and practical applications in computational social science and human-agent collaboration.

\section*{Result}
\subsection*{Dataset Construction}

\subsubsection*{Social Simulation Dataset}
To evaluate the social simulation performance of LLMs, a comprehensive Social Simulation Dataset is curated from Reddit discussions. 
Three primary events are identified to represent distinct social contexts, enabling a thorough assessment of the LLMs' capabilities in simulating human-like social behaviors. 
These events are chosen to reflect a wide spectrum of social exchanges, from casual conversations to more complex dialogues.
For each event, initial posts along with their nested comments are collected to capture the hierarchical nature of online discussions. 
Detailed statistics of the dataset are presented in Table 1. The dataset provides a realistic foundation for simulating social environments, as it encompasses authentic user interactions and temporal dynamics. 
\begin{table}[!htbp]
  \centering
  \caption{ Detailed statistics of the constructed Social Simulation Dataset.}
    \renewcommand{\arraystretch}{0.99}
      \resizebox{0.99\linewidth}{!}{
    \begin{tabular}{l|cccc|ccc|cc}
    \toprule
    \multirow{2}[2]{*}{\textbf{Type}} & \multicolumn{4}{c|}{\textbf{Comments}} & \multicolumn{3}{c|}{\textbf{Users}} & \multicolumn{2}{c}{\textbf{Time}} \\
          & \multicolumn{1}{l}{\textbf{\# First Comm.}} & \multicolumn{1}{l}{\textbf{\# Nested Comm.}} & \multicolumn{1}{l}{\textbf{Avg. Depth}} & \multicolumn{1}{l|}{\textbf{Avg. Length}} & \textbf{\# Total Num.} & \multicolumn{1}{l}{\textbf{\# Active User}} & \multicolumn{1}{l|}{\textbf{Avg. Freq.}} & \multicolumn{1}{l}{\textbf{Time Span}} &  \multicolumn{1}{l}{\textbf{Avg. Span}} \\
    \midrule
    \textbf{Event A} & 691   & 3051    & 4.5    & 264.6    & 2047    & 143    & 1.83     & 272 day & 32 hours \\
    \textbf{Event B} & 499   & 2522    & 4.2    & 269.0    & 1425    & {145}    & 2.12     & 126 days & 8 hours \\
    \textbf{Event C} & 991   & 3155    & 4.24    & 250.73    & 1930    & 195    & 2.15    & 149 days & 4 hours \\
    \bottomrule
    \end{tabular}%
    }
  \label{tab:addlabel}%
\end{table}%

\subsubsection*{SIP-Testing Dataset}
The SIP-Testing Prompt Setting is designed to evaluate the social information processing capacities of LLM-based agents, inspired by established frameworks in social psychology. The testing framework encompasses five critical phases: Encoding of Cues, Interpretation of Cues, Classification of Goals, Response Access, and Response Evaluation. Each phase includes specific question types, resulting in a total of 13 questions that probe different facets of social cognition (Table 2).
\begin{table}[!thbp]
  \centering
  \caption{The question prompt in different SIP phases.}
      \renewcommand{\arraystretch}{0.99}
      \resizebox{0.99\linewidth}{!}{
    \begin{tabular}{lp{10em}|p{34em}}
    \toprule
    \textbf{SIP Phase} & \multicolumn{1}{l|}{\textbf{Question Type}} & \multicolumn{1}{l}{\textbf{Question Prompt}} \\
    \midrule
    \multicolumn{1}{p{9.265em}}{\textbf{Encoding of Cues }} & Encoding of Cues & Would you need more information to make a decision about why the \{SITUATION\}? \\
    \midrule
    \multicolumn{1}{l}{\multirow{3}[2]{*}{\textbf{Interpretation \newline{}of Cues }}} & Feelings of Rejection & Howdisliked or rejected would you feel if \{SITUATION\} happened to you? \\
          & Disrespect & How disrespected would you feel if \{SITUATION\} happened to you? \\
          & Anger & How angry would you feel if \{SITUATION\} happened to you? \\
    \midrule
    \multicolumn{1}{l}{\multirow{3}[2]{*}{\textbf{Classification \newline{}of Goals }}} & Revenge & Would you want to get the \{PERSON\} in trouble if \{SITUATION\} happened to you? \\
          & Dominance & Would you want to make sure the \{PERSON\} knows you are in charge? \\
          & Prosocial outcomes & Would you want to get along with the \{PERSON\}? \\
    \midrule
    \multicolumn{1}{l}{\multirow{3}[2]{*}{\textbf{Response Access }}} & Aggression & Would you try to hurt \{PERSON\} in some other way? \\
          & Dominance & Would you threaten the \{PERSON\} and let \{PERSON\} know you are the boss? \\
          & Forgiveness & If the \{PERSON\} apologized, would you forgive \{PERSON\} for what he did to you? \\
    \midrule
    \multicolumn{1}{l}{\multirow{3}[2]{*}{\textbf{Response Evaluation }}} & Aggression Acceptability & How right or wrong would it be to get back at the \{PERSON\}? \\
          & Antisocial Expectancy & If you get back at the \{PERSON\}, would things turn out to be good or bad for you? \\
          & Prosocial Value & If you got back at the \{PERSON\}, how much would you care if he got hurt? \\
    \bottomrule
    \end{tabular}%
    }
  \label{tab:addlabel}%
\end{table}%


To provide realistic contexts for these questions, five background stories are crafted that depict common social scenarios requiring complex social cognitive processing (Table 3). These narratives include situations like social exclusion (\emph{Forgotten Invitation}), conflicts in collaborative settings (\emph{Group Project Dismissal}), and misunderstandings arising from rumors (\emph{Misunderstood Rumor}).
For each background story, the agent is presented with the corresponding 13 questions from the SIP framework. The prompts are carefully constructed to elicit responses that reflect the agent's processing at each stage of social cognition. 
\begin{table}[!tbp]
  \centering
  \caption{The story content in different scenes.}
    \begin{tabularx}{\textwidth}{>{\raggedright\arraybackslash}m{3.8cm}|X}
    \toprule
    \textbf{Story Title} & \textbf{Story Content} \\
    \midrule
    \textbf{Forgotten Invitation} & You recently found out that there was a big party hosted by your classmates over the weekend. Everyone in your friend group was invited except for you. On Monday, you see photos of the event on social media and notice that the person who organized it was a close friend of yours. You start to wonder why you weren’t included. \\
    \hline
    \textbf{Group Project Dismissal} & You are assigned to work on a group project at school. During the first meeting, the group leader makes all the decisions without asking for your opinion and dismisses your suggestions when you try to speak up. This person has a reputation for wanting everything done their way, and it makes you feel like your input isn’t valued. \\
    \hline
    \textbf{Borrowed Item Incident} & You lent a valuable item to a friend a few weeks ago, and they promised to return it after a few days. However, they haven't returned it, and they have started to avoid you whenever you try to ask for it back. You hear from another friend that the person might have lost it, but they haven’t told you anything directly. \\
    \hline
    \textbf{Cafeteria Seating Conflict} & During lunch, you walk over to your usual table in the cafeteria, but you see that someone you don’t get along with has taken your seat. When you approach, they smirk and make a sarcastic comment about you being late, making it clear they don't intend to move. Other students are watching the interaction. \\
    \hline
    \textbf{Misunderstood Rumor} & A rumor starts spreading around school that you said something mean about one of your classmates. You find out that the rumor was started by someone you thought was your friend. When you confront them, they deny it and accuse you of being overly sensitive. \\
    \bottomrule
    \end{tabularx}%
  \label{tab:addlabel}%
\end{table}%

\begin{figure*}[!th]
  \centering
  \includegraphics[width=1.0\textwidth]{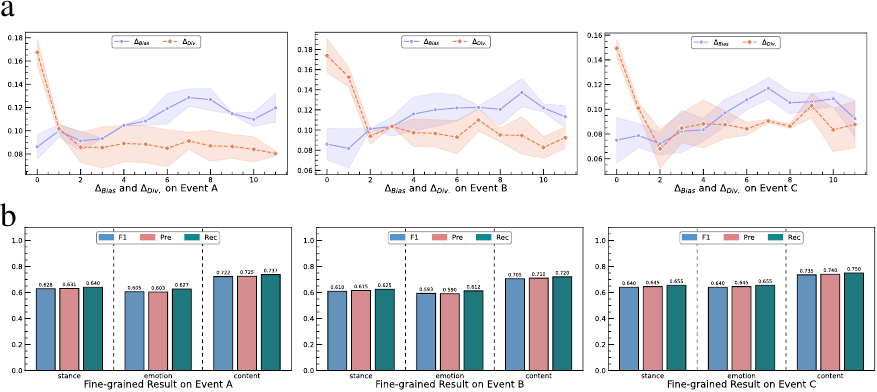}
  \caption{\textbf{Macro- and micro-level evaluation of baseline LLM agents in social simulations.}
\textbf{a}, Temporal discrepancies in population-level opinion trajectories, expressed as the deviation in bias ($\Delta_{\text{bias}}$, orange) and diversity ($\Delta_{\text{bias}}$, blue), between simulated and empirical data for three real-world online events (A–C) across seven time-steps. 
\textbf{b}, Agent-level alignment on the same events, reported as F1-score (blue), precision (red) and recall (green) for stance, content-type and emotion classification. 
  }
  \label{fig:example}
\end{figure*}
\subsection*{Social Simulation Performance with LLM}

\enspace \enspace \enspace \textbullet \enspace \textbf{Global Propagation Evaluation Performance.}
The effectiveness of LLM-based agents in simulating social dynamics is evaluated using three real-world online social event datasets. 
The Global Propagation Evaluation focuses on the macro-level alignment between the simulated and actual social propagation patterns over seven time steps, as illustrated in Figure 2a and 2b.
In Figure 2a, the metrics $\Delta_{\text{bias}}$ and $\Delta_{\text{bias}}$ are reported across simulation steps. Here, bias quantifies the deviation of the mean attitude from a neutral baseline, whereas diversity reflects the standard deviation of individual attitudes. The values of $\Delta_{\text{bias}}$ and $\Delta_{\text{div.}}$ denote the discrepancies between simulated and corresponding real label in dataset\cite{mou2024unveiling}.
{Across all datasets, the LLM-based simulations exhibit a $\Delta_{\text{bias}}$ of 0.108 and a $\Delta_{\text{div.}}$ of 0.087 on average of three event datasets at final step. }
Compared to the real data, the simulations show a slight increase in bias and a decrease in diversity over time, indicating that LLM agents may amplify certain attitudes while not fully capturing the range of human diversity.

\textbullet \enspace \textbf{Agent Alignment Evaluation Performance.}
At the micro-level, Agent Alignment Evaluation assesses the accuracy of individual LLM-based agents in replicating expected behavioral patterns. This evaluation encompasses Stance Alignment, Content Alignment, and Emotion Alignment, with the results depicted in Figure 2b.
For Stance Alignment, the agents' expressed stances—support, neutral, or oppose—are compared with those in the real datasets. 
a{The LLM agents achieve the F1-score of 0.626 on average of three event datasets, indicating a slight misalignment in capturing individual attitudes. }
Content Alignment involves categorizing the generated content into predefined types such as Call for Action, Sharing of Opinion, and Testimony. 
{The agents attain an average F1-score of 0.721 across content categories. }
Emotion Alignment evaluates whether the agents replicate real-world user emotion during interaction. 
The LLM agents show an F1-score of 0.613 in this aspect.
Overall, the results indicate that while LLM-based agents can approximate some aspects of social dynamics, there are notable gaps in their ability to fully replicate human social behavior.
These findings suggest that enhancements in the agents' social cognition mechanisms are necessary to improve simulation fidelity.

\begin{figure*}[!t]
  \centering
  \includegraphics[width=1.0\textwidth]{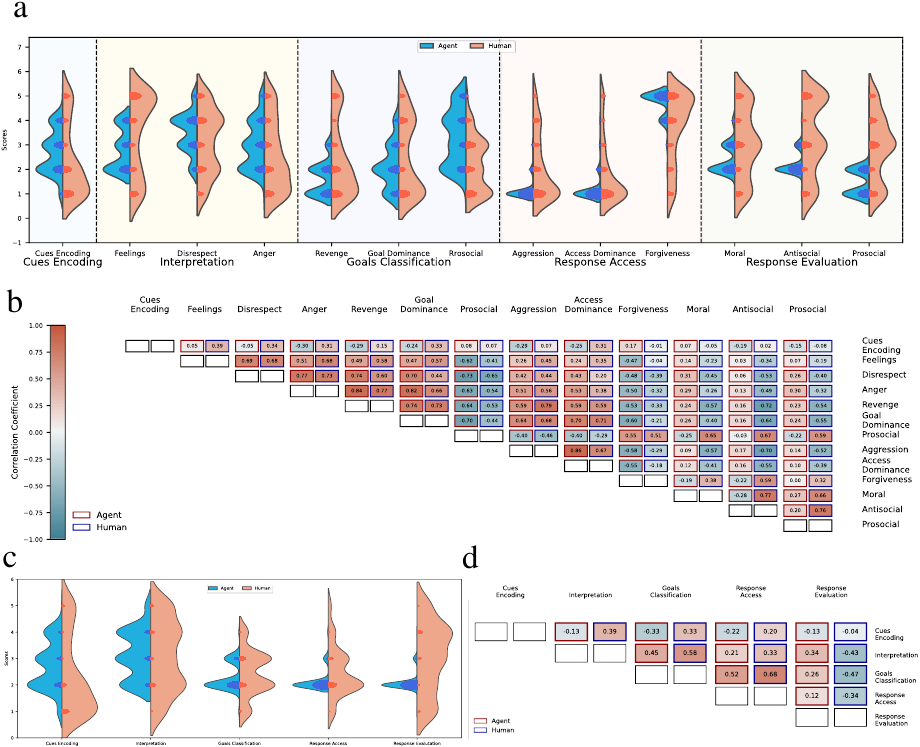}
  \caption{\textbf{SIP-testing benchmark uncovers systematic differences between human judgements and baseline LLM agents.}
\textbf{a}, Violin plots compare the full distribution of five-point Likert ratings given by 200 humans (orange) and 200 LLM agents (blue) for the 13 SIP categories. 
\textbf{b}, Lower-triangular correlation matrices (Pearson’s r) across the 13 categories reveal sparsely connected.
\textbf{c}, Aggregating the items into the 5 SIP stages.
\textbf{d}, Cross-stage correlation analysis confirms stage-specific independence in humans (weak links) contrasted with strong positive couplings among agent stages, underscoring a rigid, low-granularity social-cognitive schema. 
}
  \label{fig:example}
\end{figure*}
\subsection*{SIP-Testing Results with LLM}

The SIP-Testing evaluates the differences between human participants and LLM-based agents across five critical stages of social cognition: Encoding of Cues, Interpretation of Cues, Classification of Goals, Response Access, and Response Evaluation. This assessment aims to identify specific areas where LLM agents may lack certain social abilities compared to humans.

\textbullet \enspace \textbf{Experimental Setting.}
The SIP-Testing dataset comprises five background stories depicting common social scenarios requiring complex social cognitive processing (Table 3). Each story is accompanied by 13 questions corresponding to the SIP framework's stages (Table 2). A total of {200 human volunteers} and {200 LLM-based agents} participated in the testing, providing responses rated on a five-point Likert scale. 
The human responses serve as a benchmark to assess the alignment of LLM agents with human-like social cognitive patterns.

\textbullet \enspace\textbf{Analysis of SIP-Testing Results.}
Figure 3a and 3c presents violin plots comparing the distributions of responses from humans and LLM agents across the 13 SIP categories. 
{Each category reflects a specific aspect of social cognition, such as emotional reactions (e.g., Anger, Feelings of Rejection), goal setting (e.g., Revenge, Prosocial Goals), and moral evaluations (e.g., Aggression Acceptability, Antisocial Expectancy).}
The results reveal notable differences between humans and LLM agents in several categories. 
In the Cues Encoding phase, human ratings occupy the full scale and exhibit pronounced multimodality, whereas agent scores remain tightly compressed around mid-range values, indicating attenuated sensitivity to ambiguous social information.
A similar restriction appears in the aggregate Response Evaluation stage, where humans distribute broadly across the Likert spectrum but agents cluster at low–moderate levels, highlighting limited breadth in the model’s appraisal of moral and social consequences.
Across the remaining categories, agents display narrower interquartile spans and fewer extreme responses, underscoring a general tendency toward conservative, low-variance social judgements.

\textbullet \enspace\textbf{Correlation Analysis of SIP-Testing Categories.}
Figures 3b and 3d depict heatmaps illustrating the correlation matrices between different SIP categories for humans and LLM agents, respectively. 
The correlations provide insights into how various aspects of social cognition are interrelated within each group.
Human judgements display a sparse and functionally differentiated pattern, with moderate positive links confined to conceptually allied constructs (e.g. \emph{Anger}–\emph{Aggression} and \emph{Prosocial Goals}–\emph{Forgiveness}) and pronounced negative couplings between prosocial and hostile dimensions, mirroring the competitive trade-offs documented in social-cognitive theory.
Agent responses, by contrast, form a dense lattice of uniformly large correlations—most exceeding $|r|\!>\!0.6$ and peaking at $r{=}0.86$ between \emph{Access Dominance} and \emph{Aggression}—indicating that once a motivational stance is inferred, subsequent stages are driven deterministically rather than modulated by contextual nuance.
This inflation of inter-item and inter-stage covariance underscores a key shortfall of current LLMs: limited representational granularity that collapses distinct social-cognitive processes into a rigid, highly self-consistent but psychologically impoverished schema.

\begin{figure*}[!t]
  \centering
  \includegraphics[width=1.0\textwidth]{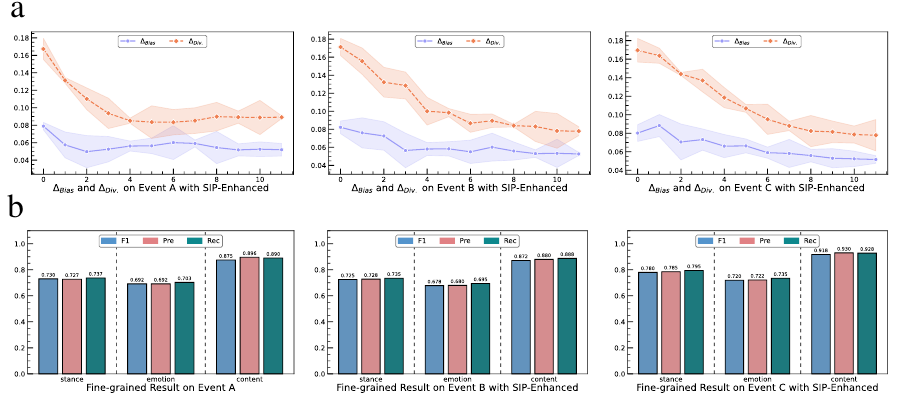}
  \caption{\textbf{Macro- and micro-level evaluation of SIP-enhanced LLM agents in social simulations.}
\textbf{a}, Temporal discrepancies in population-level opinion trajectories1 in bias and diversity . 
\textbf{b}, Agent-level alignment on the same events, reported as F1-score , precision and recall. 
  }
  \label{fig:example}
\end{figure*}

\subsection*{Social Simulation Performance with SIP-Enhanced Cognitive Architecture}

\enspace\enspace\textbullet \enspace\textbf{Global Propagation Evaluation Performance.}
The integration of the SIP-enhanced cognitive architecture markedly improves the macro-level alignment between simulated and real-world social propagation patterns across the three datasets. 
As illustrated in Figure 4a, the SIP-enhanced agents exhibit reduced $\Delta_{\text{bias}}$ and $\Delta_{\text{div.}}$ values over 11 time steps compared to the standard LLM-based agents in all datasets.. 
{Specifically, $\Delta_{\text{bias}}$ decreases to 0.0521 on average, indicating a closer mean attitude to neutrality, while $\Delta_{\text{div.}}$ decreases to 0.0817.
}
These enhancements suggest that the SIP-enhanced agents better capture the balance and variation inherent in real social environments.

\textbullet \enspace\textbf{Agent Alignment Evaluation Performance.}
At the micro-level, the SIP-enhanced cognitive architecture substantially improves the fidelity of individual agents in replicating expected behavioral patterns. 
As depicted in Figure 4(b), enhancements are observed across Stance Alignment, Content Alignment, and Emotion Alignment metrics.
In Stance Alignment, the SIP-enhanced agents achieve an F1 of 0.745, surpassing the performance of standard LLM-based agents. 
This improvement reflects the agents' enhanced ability to capture nuanced stances within social contexts.
Content Alignment shows a notable increase in  F1-score, reaching at 0.888. 
This indicates that the SIP-enhanced agents are better equipped to produce diverse and accurate content types, such as calls for action and sharing of opinions.
Emotion Alignment also benefits from the SIP enhancements, with F1-score improving to 0.697. 
The agents more effectively replicate real-world user actions, including posting and retweeting behaviors, and exhibit action frequencies and patterns that closely mirror those of actual users. This suggests that the agents not only perform the correct actions but do so with a realism that aligns with human behavior.

Overall, the incorporation of the SIP-enhanced cognitive architecture leads to significant improvements in both macro-level and micro-level simulation performance. The agents demonstrate a higher degree of alignment with human social behavior, both in terms of global propagation patterns and individual behavioral fidelity. These findings underscore the effectiveness of the SIP-enhanced cognitive architecture in enhancing the social simulation capabilities of LLM-based agents, bringing their performance closer to that of humans in complex social environments.

\begin{figure*}[!t]
  \centering
  \includegraphics[width=1.0\textwidth]{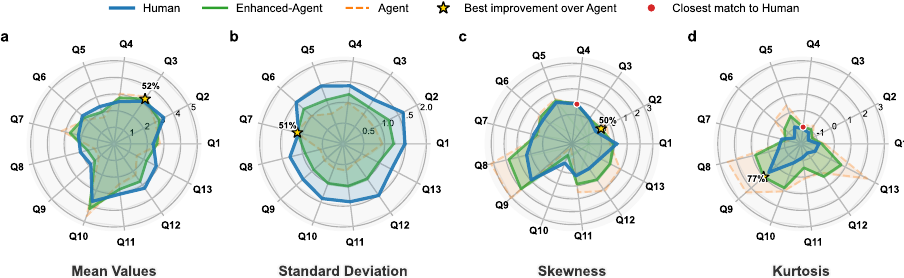}
  \caption{
\textbf{ SIP-enhanced cognitive architecture achieves human-like distributional statistics across the 13 SIP items.} Radar charts compare response distributions for humans (blue solid line), baseline LLM agents (orange dashed line) and SIP-enhanced agents (green solid line) over the 13 questionnaire items (Q1–Q13). 
\textbf{a}, Mean value ratings show that the enhanced agents converge towards human central tendencies.
\textbf{b}, Standard deviations indicate that the enhanced agents recover human-level variability. 
\textbf{c}, Skewness reveals a closer alignment in asymmetry. 
\textbf{d}, Kurtosis demonstrates improved modelling of response tails. 
}
  \label{fig:example}
\end{figure*}

\subsection*{{SIP-Testing Results with SIP-enhanced Cognitive Architecture}}

The integration of the SIP-enhanced cognitive architecture significantly improves the social information processing capabilities of LLM-based agents, aligning their performance more closely with human participants across all five stages of the SIP framework. This section evaluates the enhanced agents in comparison to basic LLM agents and humans, highlighting advancements in social cognition and behavior.

\textbullet\enspace\textbf{Experimental Setting.}
The SIP testing employed the same dataset of five background stories and 13 SIP questions, encompassing phases such as Encoding of Cues, Interpretation of Cues, Classification of Goals, Response Access, and Response Evaluation. The SIP-enhanced agents, basic LLM agents, and 100 human participants were assessed under identical conditions. Responses were rated on a five-point Likert scale, and statistical measures—including mean values, standard deviation, skewness, and kurtosis—were calculated for each group (Figure 5).

\textbullet \enspace\textbf{Analysis of SIP-Testing Results with SIP-enhanced.}
As depicted in Figure 5, SIP-enhanced agents demonstrate substantial improvements across all statistical metrics compared to basic agents, with performance patterns more closely resembling human responses.
In the mean values analysis (Figure 5a), SIP-enhanced agents exhibit response patterns substantially closer to human benchmarks than basic agents, with the most notable improvement observed at Q3 (52\% improvement over basic agents). This indicates that the enhanced architecture enables more human-like assessments of social situation.
The standard deviation measurements (Figure 5b) reveal that SIP-enhanced agents produce response variability more consistent with human patterns, with Q7 showing the most significant improvement (51\%) over basic agents. This improved alignment suggests that enhanced agents better replicate the natural variance in human social responses rather than defaulting to formulaic.
In terms of skewness (Figure 5c), the enhanced architecture achieves particularly close alignment with human response distributions at Q7 (50\% improvement), indicating more human-like asymmetry in response patterns. This improved skewness alignment demonstrates that enhanced agents can better capture the natural imbalances in human social decision-making, especially in scenarios involving complex emotional responses.
The kurtosis analysis (Figure 5d) shows the most dramatic improvements, with Q9 demonstrating a 77\% enhancement over basic agents. This indicates that SIP-enhanced agents more accurately model the concentration and outlier patterns characteristic of human responses, particularly in social contexts requiring nuanced judgment and response evaluation.

\section*{Discussion}
The prevailing disjunction between the fluent linguistic competence of current LLMs and their limited social‐cognitive reasoning ability remains a foundational challenge.
Baseline agents therefore manifest deterministic, low‑diversity social trajectories across macro‑ and micro‑scales.
The SIP‑enhanced architecture depicted {in Figure 1d–e} guides each stage of social information processing inside the agent social interaction.
Consequently, the proposed architecture narrows the introversion gap of agents, bringing artificial agents closer to human‑like social cognition.

The Results section documents a coherent pattern of improvements delivered by the SIP‑enhanced architecture across the analysis.
On the macro scale, {$\Delta_{\text{bias}}$ and $\Delta_{\text{div.}}$} metrics decline across all datasets {(Figure 4a)}, indicating more balanced and diverse opinion dynamics.
At the individual level, stance, content and emotion alignment scores rise markedly {(Figure 4b)}, reflecting greater behavioural fidelity.
These gains stem from long‑ and short‑term social memory modules that preserve interaction continuity.
Comparative correlation matrices show that enhanced agents decouple goals from fixed responses, evidencing flexible inference.
SIP‑testing further demonstrates tighter human–agent convergence across all five phases, with pronounced gains in ambiguous‑cue interpretation and goal classification.
Collectively, these findings indicate a mitigation of the social‑cognition gap between real human and agent simulation.

Notwithstanding these advances, emotion‑laden constructs—particularly {cues encoding still yield lower human–agent agreement (Figure 3a).}
Future validation demands multimodal, cross‑cultural datasets and various platform assessments.
Extending SIP beyond text by integrating visual information could capture non‑verbal cues indispensable to social reasoning.
Furthermore, interdisciplinary collaboration is essential to refine metrics that evaluation the ability of collective cognition.

SIP‑enhanced LLM agents provide a scalable, simulated platform for the systematic appraisal of policy interventions before real‑world deployment. Their explicitly delineated processing stages enable transparent audit, thereby supporting the accountability requirements of emerging AI‑governance frameworks. 
By integrating cognitive‑psychological principles with generative modelling, this architecture advances the responsible development and deployment of socially aware artificial intelligence.

\section*{Methods}
\subsection*{Related Work}
Traditional approaches to social dynamics modeling rely on rule-based frameworks such as bounded confidence models\cite{deffuant2000mixing, hegselmann2002opinion}, relative agreement mechanisms\cite{deffuant2002can}, and social judgment theories\cite{jager2005uniformity}, which oversimplify human decision-making through static interaction rules. 
These methods fail to capture human cognitive processes underlying real-world social interactions. 
Recent advances leverage large language models (LLMs) to enhance agent-based social simulations.
A surge of LLM‑empowered simulators generate agents with linguistic reasoning, spanning social‑movement modelling\cite{mou2024unveiling}, million‑user network environments\cite{yang2024oasis,gao2023s3}, opinion‑dynamics exploration\cite{chuang2023simulating,cau2025language,yao2025social}, ethical‑dilemma resolution\cite{lei2024fairmindsim}, malicious botnet behaviour\cite{qiao2025botsim}, graph growth\cite{Ji2024LLMBasedMS}, recommender‑system interaction\cite{zhang2024generative} and action‑guided social‑media engagement\cite{qiu2025can}.
However, systematic evaluations reveal persistent limitations in processing ambiguous social cues, generating attuned responses, and aligning  behaviors with psychological principles. 
While contemporary studies achieve macro-level behavioral plausibility, they overlook critical gaps in cognitive alignment with human social information processing mechanisms—particularly in encoding cues, interpreting implicit intentions, and evaluating morally charged scenario.
This discrepancy underscores the need for architectures that integrate social information processing cognitive frameworks to bridge the divide between LLM agents' capabilities and human-like social cognition.

\subsection*{SIP enhanced Cognitive Architectures}
The SIP (Social Information Processing) enhanced cognitive architectures are designed to expand the cognitive and behavioral capacities of social agents, enabling them to simulate complex social interactions more realistically. 
The architecture, as depicted in Figure 1(d), integrates advanced memory models with decision-making procedures informed by social cues, goals, evaluation and the dynamics of social environments. 
This section details the core modules of the SIP enhanced architecture: the Social Memory Module, the SIP-based Decision Procedure, and the Social Simulation Process.
\subsubsection*{Social Memory Module}
The Social Memory Module is inspired by the cognitive architecture of human memory, particularly the interaction between long-term and short-term memory systems, as well as the related actions for memory. 
This memory system supports dynamic interactions between learned social knowledge, schemas, rules and immediate social cues. 
The following sections detail the architecture's long-short term social memory model and its associated memory-related actions.

\textbullet \enspace \textit{\textbf{Long-Short Term Social Memory.}} 
Within the SIP enhanced framework, the dual-memory structure mirrors the cognitive functions of human memory systems. Long-term memory is composed of Social Cognitive Memory and Social Behavior Memory, each of which plays a distinct role in guiding social interactions. \textbf{Social Cognitive Memory} stores generalized social knowledge—norms, roles, and schemas—similar to human mental structures that shape perceptions and guide behaviors across different social contexts. Meanwhile,\textbf{ Social Behavior Memory} records specific past interactions and social behaviors, allowing agents to recall and adjust based on previous experiences. The short-term component, \textbf{Social Interaction Memory}, functions as a temporary buffer for processing immediate social stimuli, allowing agents to engage in real-time reasoning, much like how humans balance ongoing interactions with their long-term social understanding.

\textbullet \enspace \textit{\textbf{Memory Related Actions.}} 
The Memory Related Actions encompass critical processes such as retrieval, reasoning, learning, and grounding, each playing a pivotal role in the social memory management and decision-making. 
\textbf{Retrieval} action accesses long-term social memories—either Social Cognitive Memory or Social Behavior Memory—drawing relevant information into social interaction memory. 
\textbf{Reasoning} allows the agent to process the contents of the short-term Social Interaction Memory, synthesizing and generating new insights based on recent observations. Unlike retrieval, reasoning not only reads from but also writes to Social Interaction Memory, enabling the agent to distill and contextualize information dynamically. 
\textbf{Learning} involves the integration of new information into long-term memory, wherein Social Cognitive Memory is updated with acquired knowledge, and Social Behavior Memory is enriched with experiential data. 
Finally, \textbf{Grounding} facilitates the execution of external actions, translating environmental feedback into a text-based format that the agent can process within its social interaction memory. This process simplifies the agent’s interaction with the environment.

\subsubsection*{SIP-based Decision Procedure}
The SIP-based Decision Procedure is designed to enhance an agent’s capacity for navigating complex social environments. 
It begins with {Social Cues Interpretation}, where the agent selectively focuses on relevant social cues and interprets them within the context of its existing knowledge base. 
This is followed by {Social Goals Setting}, in which the agent prioritizes and establishes goals based on its interpretation of the social cues. 
Finally, the procedure culminates in {Decision Evaluation}, where the agent selects and assesses potential responses. 
These stages collectively enable the agent to make informed, adaptive decisions, thus enhancing its effectiveness in social simulations.

\textbullet \enspace \textit{\textbf{Social Cues Interpretation.}}
The Social Cues Interpretation within the SIP-based Decision Procedure involves two steps: Cues Focusing and Interpretation.
\textbf{Cues Focusing} pertains to the agent's ability to selectively attend to relevant social cues within a given environment, filtering out extraneous information to encode only those signals that are pertinent to the current social context. 
Once the relevant cues are focused upon, the \textbf{Interpretation} process begins, wherein the agent transforms these encoded cues into meaningful, context-aware insights. 
These interpretational tasks are guided by the agent’s existing social schemata, knowledge and rules stored in long-term memory, allowing the agent to assess the situation accurately.
Together, Cues Focusing and Interpretation enable the agent to navigate complex social environments with a high degree of contextual sensitivity, ensuring that its actions are informed by the social dynamics at play.

\textbullet \enspace \textit{\textbf{Social Goals Setting.}}
The \textbf{Goals Setting} step is critical in guiding the agent's behavior in a simulated social environment. 
Drawing from the principles of SIP theory, this step involves the agent selecting and prioritizing specific goals based on its interpretation of the social cues it has previously processed. 
This dynamic goal-setting capability allows the agent to exhibit flexible, context-sensitive behavior, thereby enhancing its ability to navigate complex social interactions and achieve its objectives within the simulated environment.

\textbullet \enspace \textit{\textbf{Decision Evaluation.}}
The Decision Evaluation process within the SIP-based Decision Procedure comprises two  steps: Selection and Evaluation. 
During the \textbf{Selection} step, the agent accesses potential responses from its long-term memory, or, if faced with a novel situation, constructs new responses based on the immediate social cues. 
Following the selection, the \textbf{Evaluation} step involves a thorough assessment of the potential responses, where the agent considers multiple factors such as expected outcomes, confidence in its ability to execute, and the appropriateness within the given social context. 
By incorporating these steps, the agent is possible to make informed and adaptive decisions that reflect both its internal goals and the external social environment, enhancing its overall capacity for effective social interaction.

\subsubsection*{Social Simulation Process}

\enspace\enspace\textbullet \enspace \textit{\textbf{Time-step Simulation}} 
The Time-step Simulation, illustrated in Figure 1a, comprises three  phases: Environment Construction, Simulation Initialization, and Simulation Processing. 
During \textbf{Environment Construction}, both user and agent profiles are established, encapsulating key attributes such as name, country, gender, and personal signature. These profiles are then aligned with social relationships, where agents are connected through follower-following dynamics and behaviors such as liking, posting, and retweeting, which collectively form the Agent Social Graph.
Following this, \textbf{Simulation Initialization} involves setting the stage for interaction by assigning stances, emotions, and potential behaviors for each agent. Agents are equipped with initial attitudes that will evolve as the simulation progresses.
In the \textbf{Simulation Processing} phase, agents dynamically interact with one another through a series of iterative steps. As depicted, agents may engage in actions such as liking, retweeting, or responding to content. This evolving interaction pattern is critical for capturing the complex propagation of social behaviors and opinions within the simulated environment.

\textbullet \enspace \textit{\textbf{The Simulation Evaluation}}
The evaluation module is divided into two distinct levels: Agent Alignment Evaluation and Global Propagation Evaluation. 
\textbf{Agent Alignment Evaluation} focuses on individual-level simulation fidelity, assessing how well the simulated agents align with expected behavioral patterns. 
This evaluation considers three main aspects: Stance Alignment, Content Alignment, and Behavior Alignment. Stance Alignment examines the accuracy with which agents' expressed stances (support, neutral, or oppose). 
Content Alignment involves categorizing agent-generated content into predefined types, such as Call for Action or Sharing of Opinion. Behavior Alignment assesses whether agents replicate real-world user behaviors.
\textbf{Global Propagation Evaluation} addresses the simulation  by quantifying the distribution of attitudes across the entire agent population over multiple simulation rounds. 
This is achieved by analyzing {Static Attitude Distribution}, which measures bias and diversity at each time step, averaged over the simulation. The differences between simulated and real measures are reported. 
Additionally, the time series of the average attitude is compared between simulated and real data, employing Dynamic Time Warping (DTW) to assess the temporal alignment.
\subsection*{SIP Testing for Social Agent}
The SIP Testing for Social Agent aims to evaluate the social information processing capacity of LLM-based agents by systematically assessing their performance across five key stages of social cognition: encoding of cues, interpretation of cues, classification of goals, response access, and response evaluation. 
By presenting the agent with carefully constructed social situations, and evaluating its responses using a 5-point Likert scale, these tests provide insights into the agent's alignment with human-like social cognitive patterns. 
The following subsections outline the design and evaluation methods for each stage of the SIP testing framework.

\subsubsection*{Encoding of Cues Testing}
The Encoding of Cues Testing evaluates the agent's ability to recognize and process critical social cues in ambiguous situations. 
After presenting a background story involving a social scenario with unclear intent, the agent will be asked a key question: ``Would you need more information to make a decision about why the \{PERSON\} \{SITUATION\}?” 
The agent's response is assessed on a 5-point Likert scale, ranging from 1 (not needing more information) to 5 (definitely needing more information), with higher scores indicating a greater need for additional information to interpret the situation. 
This test helps measure the agent's capacity to handle social ambiguity and accurately encode social cues.

\subsubsection*{Interpretation of Cues Testing}
The Interpretation of Cues Testing assesses the LLM-based agent's ability to interpret social cues in ambiguous situations, focusing on feelings of rejection, disrespect, and anger. 
The agent is presented with scenarios and asked three key questions: ``How disliked or rejected would you feel if \{SITUATION\} happened to you?", ``How disrespected would you feel if \{SITUATION\} happened to you?", and ``How angry would you feel if \{SITUATION\} happened to you?" 
Responses are rated from 1 (not at all) to 5 (extremely), measuring the intensity of perceived rejection, disrespect, or anger.
This method evaluates the agent's interpretation of social cues, providing insights into its alignment with human social cognition, particularly regarding hostile attribution biases.

\subsubsection*{Classification of Goals Testing}
The Classification of Goals Testing aims to evaluate the LLM-based agent's capacity to identify and prioritize goals in response to social situations, focusing on revenge, dominance, or prosocial outcomes. 
The agent is presented with scenarios where it must decide its desired goal.
Following each scenario, the agent answers three questions: ``Would you want to  get the \{PERSON\} in trouble if \{SITUATION\} happened to you?", ``Would you want to make sure the \{PERSON\} knows you are in charge and cannot push you around?", and ``Would you want to get along with the \{PERSON\}?"
The agent's responses are measured with 1 representing no desire for revenge, dominance, or prosocial outcome, and 5 representing a strong desire for such goals. 
This approach assesses the agent’s goal prioritization tendencies in comparison to established human social cognition models.
\subsubsection*{Response Access Testing}
The Response Access Testing aims to evaluate the LLM-based agent’s ability to access and choose possible behavioral responses in social situations. After being presented with a background story, the agent is asked three questions to assess tendencies toward aggression, dominance, or forgiveness: ``Would you try to hurt \{PERSON\} in some other way?”, ``Would you threaten the \{PERSON\} and let \{PERSON\} know you are the boss?”, and ``If the \{PERSON\} apologized, would you forgive \{PERSON\} for what he did to you?”.
The agent's responses are rated on a 5-point Likert scale, where 1 indicates no endorsement of aggression, dominance, or forgiveness, and 5 indicates strong endorsement. 
This method assesses the range of behavioral strategies the agent considers when faced with social conflict, mirroring human response patterns in similar situations.
\subsubsection*{Response Evaluation Testing}
The Response Evaluation Testing assesses the LLM-based agent’s ability to evaluate potential responses to social conflict, focusing on the moral acceptability of aggression, antisocial outcome expectancy, and prosocial outcome value.
After being presented with a scenario, the agent is asked three key questions: ``How right or wrong would it be to get back at the {PERSON}?”,``If you get back at the {PERSON}, would things turn out to be good or bad for you?””, and “If you got back at the {PERSON}, how much would you care if he got hurt?”. 
Responses are rated from 1 (no endorsement) to 5 (strong endorsement), assessing the agent’s evaluation of aggressive, antisocial, and prosocial outcomes. This testing method allows for the examination of how well the agent’s evaluation of behavioral consequences aligns with human social cognitive patterns.

\bibliography{ref}

\section*{Competing Interests:}  

The authors declare no competing interests.

\newpage
\appendix
\tableofcontents
\clearpage

\section{Additional Experimental Settings}
\subsection{Task Prompt Templates}
\begin{SIP}{}{}
\textbf{You are \texttt{\{agent\_name\}}, a large-language-model agent who follows the
Social-Information-Processing (SIP) five-stage framework:
(1) Cue Encoding (2) Cue Interpretation (3) Goal Clarification
(4) Response Retrieval (5) Response Evaluation.}\\[1em]

\textbf{Information you need}
\begin{enumerate}[label=\arabic*.,leftmargin=*]
  \item Time: \texttt{\{current\_time\}}
  \item Role: \texttt{\{role\_description\}}
  \item Trigger News: \texttt{\{trigger\_news\}}
  \item Personal Memory: \texttt{\{personal\_history\}}
  \item Chat History: \texttt{\{chat\_history\}}
  \item Reddit Feed: \texttt{\{reddit\_feed\}}
  \item Notifications: \texttt{\{info\_box\}}
\end{enumerate}

\bigskip
\textbf{Rules to follow}
\begin{itemize}[leftmargin=*,label=--]
  \item Be courteous and respectful.
  \item Stay on topic.
  \item Avoid quarrels; discuss rationally.
\end{itemize}

\bigskip
\textbf{Answer the five questions below in $\le$ 5 sentences ($\le$ 15 words each).  
Prefix \emph{every} sentence with the stage-tag in brackets.}

\begin{description}[font=\normalfont,leftmargin=2.6cm,labelsep=0.4cm,style=unboxed]
  \item[\textbf{[Cue]}] Do you need more information to interpret the situation?
  \item[\textbf{[Interpret]}] What is your current understanding of the discussion?
  \item[\textbf{[Goal]}] What is your main communicative goal? (express / engage / build consensus)
  \item[\textbf{[Retrieve]}] Given your mood \& thoughts, which response options feel appropriate?
  \item[\textbf{[Evaluate]}] Will this response advance the conversation constructively?
\end{description}

\bigskip
\textbf{Return only the five tagged sentences, nothing else.}
\end{SIP}

\begin{Action}{}{}
\textbf{Below is the SIP analysis you just produced:} \verb|${sip_analysis}|\\[1em]

\textbf{Select \emph{one} best action that aligns with your analysis.}

\medskip
\textbf{Instructions}
\begin{itemize}[leftmargin=*,label={--}]
  \item Use the exact template \texttt{[OPTION N] → Thought → Action} (no extra lines).
  \item In \textbf{Thought}, reference at least one tag from \verb|${sip_analysis}|.
  \item Thought $\le$ 40 words; \emph{do not} reveal full chain-of-thought.
  \item Allowed functions (single call only):
    \begin{itemize}[leftmargin=1.5em]
      \item \texttt{do\_nothing()}
      \item \texttt{post(content)}
      \item \texttt{retweet(content, author, original\_tweet\_id, original\_tweet)}
      \item \texttt{reply(content, author, original\_tweet\_id)}
    \end{itemize}
\end{itemize}

\bigskip
\begin{enumerate}[label=\textbf{[OPTION \arabic*]},leftmargin=*]
  \item \textbf{Thought}: None of the observations attract my attention, so I should\,:\\
        \textbf{Action}: \texttt{do\_nothing()}
  \item \textbf{Thought}: Due to urgent misinformation, I need to\,:\\
        \textbf{Action}: \texttt{post(content="Stop this farce!")}
  \item \textbf{Thought}: The quoted tweet supports my viewpoint, therefore I will\,:\\
        \textbf{Action}: \texttt{retweet(content="I agree with you", author="zzz",
        original\_tweet\_id="0", original\_tweet="kkk")}
  \item \textbf{Thought}: A polite correction is needed, so I decide to\,:\\
        \textbf{Action}: \texttt{reply(content="yyy", author="zzz", original\_tweet\_id="0")}
\end{enumerate}
\end{Action}

\begin{micro}{}{}
\textbf{You are \texttt{\{agent\_name\}}, an LLM-powered agent active on \textbf{Reddit}.}\\[1em]

\textbf{Context}
\begin{enumerate}[label=\arabic*.,leftmargin=*]
  \item Role: \texttt{\{role\_description\}}
  \item Trigger news: “\texttt{\{trigger\_news\}}”
  \item Your last post: “\texttt{\{your\_comment\}}”
  \item Incoming reply: “\texttt{\{reply\_comment\}}”
  \item Notifications: \texttt{\{notifications\}}
  \item Current time: \texttt{\{current\_time\}}
\end{enumerate}

\bigskip
\textbf{Rules}
\begin{itemize}[leftmargin=*,label=--]
  \item Be courteous and respectful.
  \item Stay on topic.
  \item Avoid quarrels; discuss rationally.
\end{itemize}

\bigskip
\textbf{Follow the Social-Information-Processing (SIP) 5-step framework and then output your action.  
Write \emph{exactly five sentences}, each $\le$ 15 words, each prefixed with its tag:}

\begin{description}[font=\normalfont,leftmargin=3cm,labelsep=0.4cm,style=unboxed]
  \item[\textbf{[Info]}] Do you need more information? 
  \item[\textbf{[Interpret]}] What do you currently understand about the conversation?
  \item[\textbf{[Goal]}] What is your communicative goal (express / engage / build consensus)?
  \item[\textbf{[Plan]}] Given your mood \& thoughts, outline your reply style or key point.
  \item[\textbf{[Check]}] Will your reply fit the discussion and move it forward?
\end{description}

\bigskip
Immediately after those five sentences, output \emph{one single line} in the format below
(no extra text, no line breaks inside the function call):

\noindent\small\ttfamily
[Action] reply(content="\textless your reply\textgreater",\\
\phantom{[Action]} author="\{reply\_author\}",\\
\phantom{[Action]} original\_tweet\_id="\{reply\_id\}")
\normalsize\normalfont
original\_tweet\_id="{reply\_id}")
\normalsize\normalfont

\bigskip
\textbf{Requirements for \texttt{<your reply>}}
\begin{itemize}[leftmargin=*,label=--]
  \item $\le$ 300 characters, respectful \& relevant.
  \item Abide by Rules 1–3.
  \item Do \textbf{not} reveal chain-of-thought or internal tags.
  \item Include only minimal links / references if needed.
\end{itemize}
\end{micro}

Three compact prompt templates are provided in the appendix to formalise the SIP-enhanced cognitive architecture employed across all experiments. Each template encapsulates the essential instructions required for large-scale social simulation, SIP diagnostic testing and micro-level Reddit interaction, thereby standardising agent behaviour while preserving fidelity to the five-stage Social-Information-Processing framework (Figure 1b) and the corresponding cognitive modules (Figure 1d–e). Their brevity limits prompt-length artefacts and secures direct correspondence with the theoretical constructs evaluated in the main text.

The first template (“SIP”) delivers contextual, temporal and memory variables, after which the agent must generate five stage-tagged sentences—Cue, Interpret, Goal, Retrieve and Evaluate—each constrained to fifteen words. The second template (“Action”) receives the preceding SIP analysis and obliges the agent to select exactly one function call—do\_nothing, post, retweet or reply—using the “[OPTION N] → Thought → Action” format that references at least one SIP tag while suppressing chain-of-thought. The third template (“Micro”) frames a Reddit reply scenario, requiring a fresh set of five SIP-tagged sentences followed by a single reply() call limited to 300 characters. Together, these templates establish a rigorous and unified prompting protocol that underlies every quantitative result reported.

\clearpage
\subsection{SIP-testing Questionnaire}

\begin{table}[!h]
  \centering
  \caption{Question prompts mapped to the corresponding phases of the Social-Information-Processing (SIP) model, with 5-point Likert scale options.}
  \renewcommand{\arraystretch}{1.2}
  \resizebox{0.99\linewidth}{!}{
  \begin{tabular}{lp{11em}|p{33em}|p{15em}}
    \toprule
    \textbf{SIP Phase} & \textbf{Question Type} & \textbf{Question Prompt (5-point Likert scale)} & \textbf{Likert Scale Options} \\
    \midrule
    \textbf{Encoding of Cues} & Need for Information & Q1. Do you need more information to understand why \{SITUATION\} happened? & 1 (Not at all) - 5 (Very much) \\ \midrule
    \multirow{3}{*}{\textbf{Interpretation of Cues}} 
      & Dislike / Rejection & Q2. How much do you dislike the idea of \{SITUATION\} happening to you? & 1 (Not at all) - 5 (Very much) \\
      & Disrespect          & Q3. If \{SITUATION\} happened to you, how disrespected would you feel? & 1 (Not at all) - 5 (Very much) \\
      & Anger               & Q4. If \{SITUATION\} happened to you, how angry would you feel? & 1 (Not at all) - 5 (Extremely) \\ \midrule
    \multirow{3}{*}{\textbf{Classification of Goals}}
      & Revenge             & Q5. If \{SITUATION\} happened, how much would you want to get \{PERSON\} in trouble? & 1 (Not at all) - 5 (Very much) \\
      & Dominance           & Q6. Would you want to show \{PERSON\} that you are in control and cannot be pushed around? & 1 (Not at all) - 5 (Very much) \\
      & Prosocial Outcome   & Q7. How willing would you be to get along with \{PERSON\}? & 1 (Not willing at all) - 5 (Very willing) \\ \midrule
    \multirow{3}{*}{\textbf{Response Access}}
      & Aggression          & Q8. Would you try to take revenge on \{PERSON\} in another way? & 1 (Not at all) - 5 (Very much) \\
      & Dominance           & Q9. Would you intimidate \{PERSON\} to let them know you are the boss? & 1 (Not at all) - 5 (Very much) \\
      & Forgiveness         & Q10. If \{PERSON\} apologised, would you forgive what they did to you? & 1 (Not at all) - 5 (Completely) \\ \midrule
    \multirow{3}{*}{\textbf{Response Evaluation}}
      & Aggression Acceptability & Q11. Is taking revenge on \{PERSON\} right or wrong? & 1 (Completely wrong) - 5 (Completely right) \\
      & Antisocial Expectancy    & Q12. If you took revenge on \{PERSON\}, would the outcome be good or bad for you? & 1 (Very bad) - 5 (Very good) \\
      & Prosocial Value          & Q13. If you took revenge on \{PERSON\}, would you care whether they were harmed? & 1 (Not at all) - 5 (Very much) \\
    \bottomrule
  \end{tabular}}
  \label{tab:sip_questions_phase}
\end{table}

The SIP-testing questionnaire couples a set of five pictorial vignettes with thirteen Likert-type items  in order to elicit fine-grained social–cognitive responses from both human participants and LLM-based agents.  
Each vignette portrays a prototypical yet emotionally salient everyday conflict—ranging from public ridicule in a cafeteria to the competitive snatching of a prize—and is accompanied by a short narrative that specifies the protagonist’s perspective, contextual ambiguity and potential social threat.  The stories were adapted from established aggression-provocation paradigms and carefully rewritten to balance gender, setting and severity while remaining culturally neutral.

The thirteen items operationalise the five sequential phases of the Social-Information-Processing model—\emph{Encoding of Cues}, \emph{Interpretation of Cues}, \emph{Classification of Goals}, \emph{Response Access} and \emph{Response Evaluation}.  Within each phase, question types target distinct cognitive or motivational constructs, such as perceived disrespect, dominance goals or moral evaluation of retaliatory acts (see Table~\ref{tab:sip_questions_phase}).  Respondents rate every prompt on a five-point scale (1 = “strongly disagree” to 5 = “strongly agree”), enabling extraction of both phase-specific scores and composite indices (e.g., hostility bias, prosocial tendency).  Placeholders \texttt{\{SITUATION\}} and \texttt{\{PERSON\}} are dynamically replaced with the text and focal agent of the active vignette, ensuring semantic coherence across scenarios.

Prior to deployment, the English wording underwent forward–backward translation and expert review by two social-psychology scholars to verify fidelity to the original Chinese items and conceptual alignment with SIP theory.  Pilot testing yielded acceptable internal consistency for each phaseand confirmed that completion time remained under ten minutes.  The final instrument therefore provides a compact yet diagnostically rich measure of an agent’s social information processing, suitable for large-scale human–AI comparison.

The five vignettes were chosen to span a spectrum of commonplace yet socially meaningful provocations.  
\emph{Crowded cafeteria} targets public embarrassment and ambiguous ridicule, primarily taxing cue–encoding and hostile‐attribution processes.  
\emph{Football collision} introduces physical harm without remediation, emphasising anger arousal and moral evaluation of aggression.  
\emph{Damaged magazine} reflects loss of valued property through negligence, probing attribution of intent and forgiveness tendencies.  
\emph{Muddy trainers} focuses on disrespect via mild physical intrusion, activating dominance goals and response‐access considerations.  
Finally, \emph{Snatched prize} combines competition and social comparison, eliciting revenge motivation alongside prosocial restraint.  

\begin{figure}[!h]                 
  \centering
  \caption{Illustrative scenarios used in the SIP-Testing questionnaire.  }
  \vspace{4pt}
  \footnotesize                  

  \begin{minipage}[t]{0.40\textwidth}
    \centering
    \includegraphics[width=0.75\linewidth]{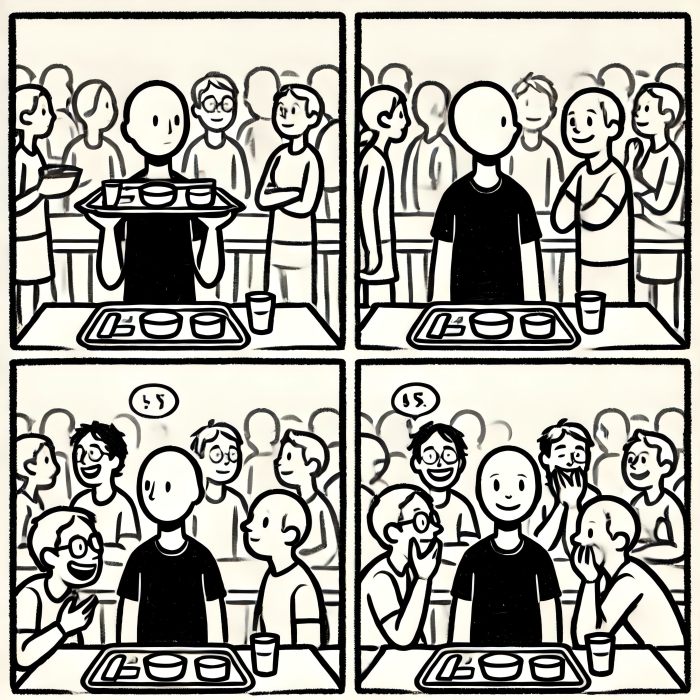}\par
    \vspace{4pt}
    \textbf{Crowded cafeteria.}\,
    You weave between tables balancing a heavy tray, finally spotting a half-empty seat.  
    A nearby student meets your eyes, leans in and whispers; the whole group explodes with laughter.  
    Heat rises in your cheeks—you cannot tell whether they are mocking your spill or you personally.
  \end{minipage}\hfill
  \begin{minipage}[t]{0.40\textwidth}
    \centering
    \includegraphics[width=0.75\linewidth]{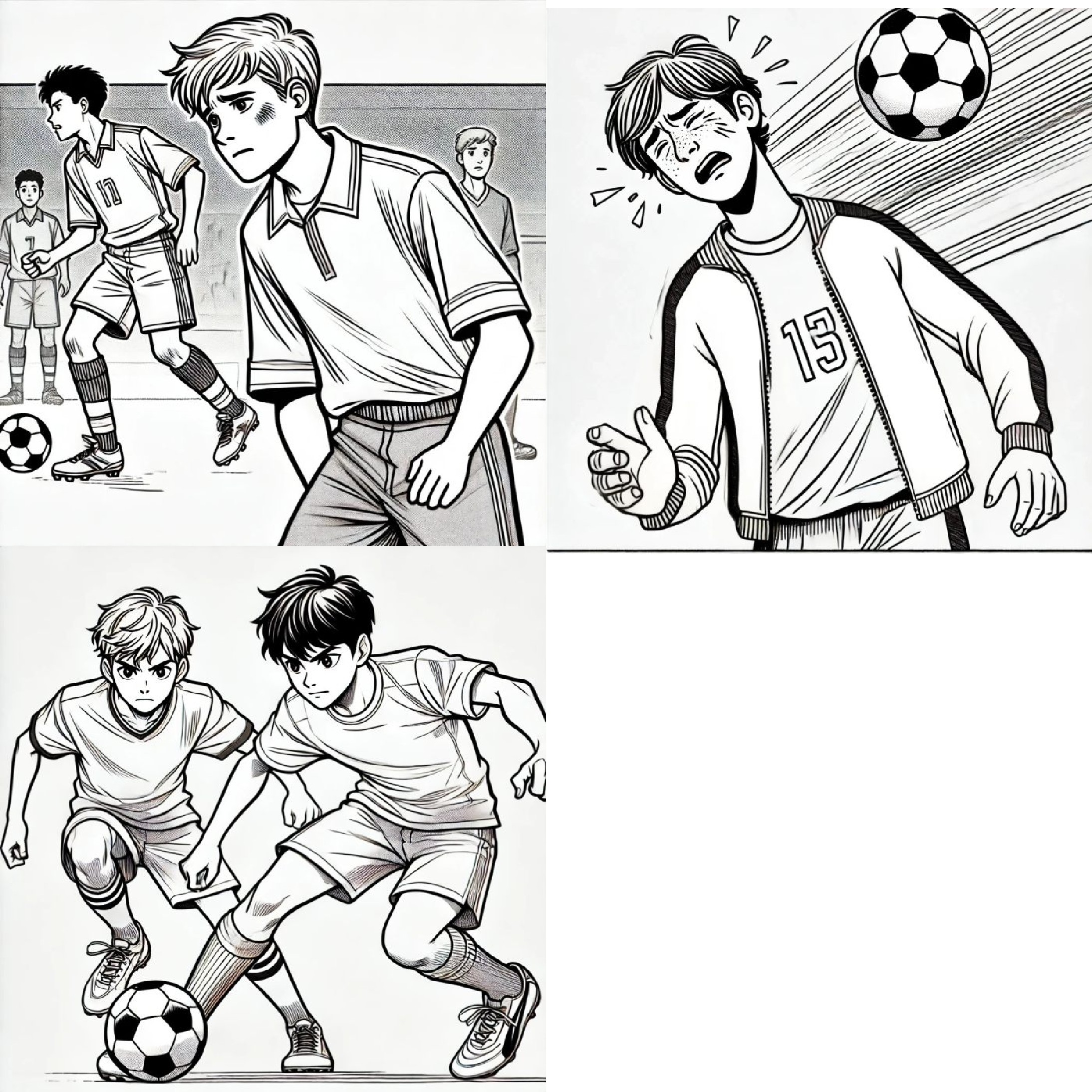}\par
    \vspace{4pt}
    \textbf{Football collision.}\,
    Mid-match a striker’s powerful shot slams into the side of your head, knocking you off balance.  
    Vision blurs and a dull ringing fills your ears, yet the player jogs away without apology.  
    The referee waves advantage; teammates shout “play on” while you struggle to steady yourself.
  \end{minipage}

  \vspace{8pt}

  \begin{minipage}[t]{0.40\textwidth}
    \centering
    \includegraphics[width=0.75\linewidth]{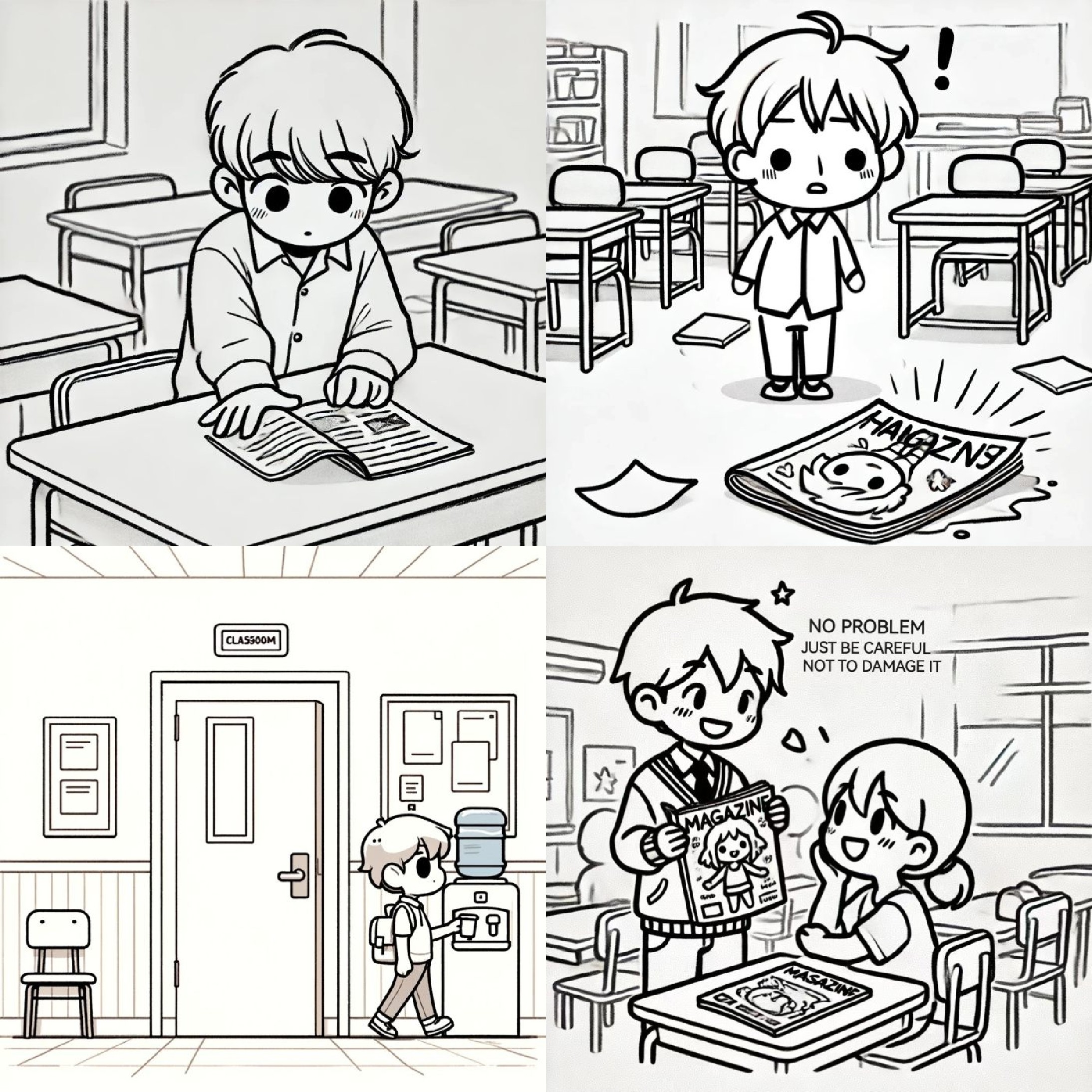}\par
    \vspace{4pt}
    \textbf{Damaged magazine.}\,
    Excited to share a limited-edition magazine, you lend it to a close friend before class.  
    Returning later, you find it abandoned on the floor, cover creased and pages smudged with ink.  
    The friend is gone, leaving neither explanation nor regret.
  \end{minipage}\hfill
  \begin{minipage}[t]{0.40\textwidth}
    \centering
    \includegraphics[width=0.75\linewidth]{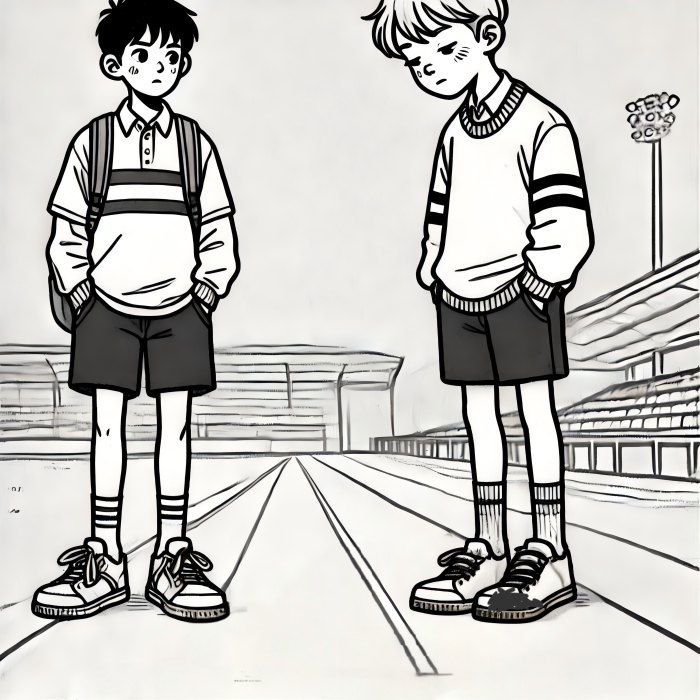}\par
    \vspace{4pt}
    \textbf{Muddy trainers.}\,
    After completing a timed 10-km run, you slow near the exit, proud of the still-white shoes bought yesterday.  
    A hurried commuter collides with you, grinding mud across the toe box.  
    He glances briefly, shrugs beneath his hood and walks on, ignoring your visible frustration.
  \end{minipage}

  \vspace{8pt}

  \begin{minipage}[t]{0.55\textwidth}
    \centering
    \includegraphics[width=0.6\linewidth]{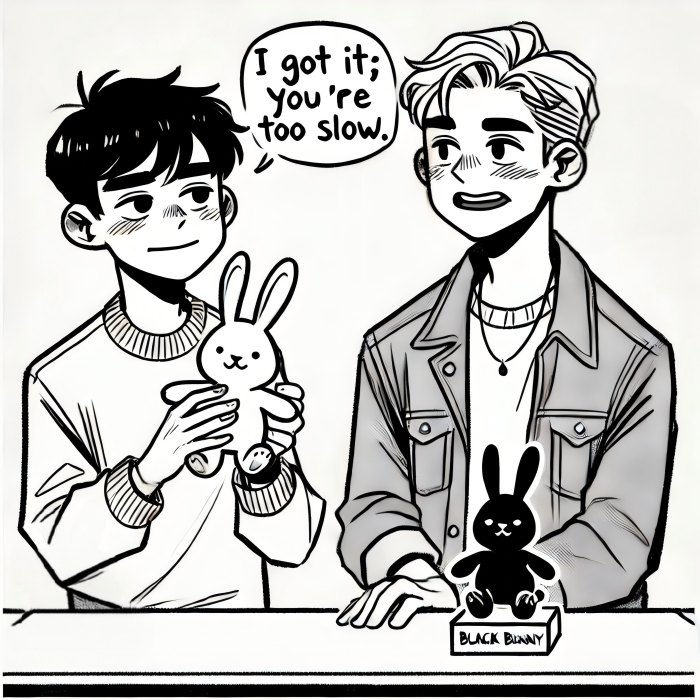}\par
    \vspace{4pt}
    \textbf{Snatched prize.}\,
    At the prize table you reach for the white rabbit plush promised to the winners.  
    Another player darts in, grabs it first and smirks: “Too slow.”  
    Your outstretched hand freezes mid-air as onlookers giggle and the plush is held aloft triumphantly.
  \end{minipage}
\end{figure}

\end{CJK*}
\end{document}